\documentclass[10pt,twocolumn]{article}
\usepackage[utf8]{inputenc}
\usepackage[T1]{fontenc}
\usepackage{mathptmx}
\usepackage[margin=0.55in,top=0.56in,bottom=0.55in,footskip=0.22in]{geometry}
\usepackage{microtype}
\usepackage{amsmath,amssymb}
\usepackage{booktabs}
\usepackage{tabularx}
\usepackage{graphicx}
\usepackage{caption}
\usepackage{enumitem}
\usepackage{hyperref}
\usepackage{url}
\usepackage{titlesec}
\usepackage{placeins}
\usepackage{flafter}
\usepackage{dblfloatfix}
\usepackage{capt-of}
\usepackage{paracol}
\usepackage{natbib}
\usepackage{etoolbox}
\bibpunct{(}{)}{;}{a}{,}{}
\AtBeginEnvironment{thebibliography}{\footnotesize\setlength{\itemsep}{0pt}\setlength{\parsep}{0pt plus 0.5pt}}

\hypersetup{colorlinks=false}
\setlist{nosep,leftmargin=*}
\titleformat{\section}{\normalfont\large\bfseries}{\thesection}{0.75em}{}
\titleformat{\subsection}{\normalfont\normalsize\bfseries}{\thesubsection}{0.75em}{}
\titlespacing*{\section}{0pt}{8pt plus 2pt}{3pt}
\titlespacing*{\subsection}{0pt}{3pt plus 1pt}{1pt}
\title{\textbf{\large The Digital Apprentice: A Framework for Human-Directed Agentic AI Development}\\[0.1em]
{\normalsize Earned Autonomy Through Observational Learning and Inference-Time Decision Memory}}
\author{Travis Weber\quad and\quad Rohit Taneja\\
{\normalsize Pheo Inc \quad|\quad \texttt{\{travis,rohit\}@pheo.ai}}}
\date{}

\begin{document}
\maketitle
\vspace{-12pt}

\begin{abstract}
\noindent
Agentic AI deployments face a recurring design tension: heavy human oversight limits scale, while broad autonomy outruns accountability. Neither posture provides the governance infrastructure required for responsible delegation.

We present the Digital Apprentice, a framework for scalable, safe AI agency in which autonomy is earned, not assumed. The Digital Apprentice is a developmental learner that internalizes the tacit methodology of a directing human, graduating through per-skill autonomy tiers only when empirical evidence justifies it. The result is an agent that becomes genuinely useful over time while remaining aligned to a specific human's standards.

Three architectural components make this possible. (1)~Methodology capture, distilling a directing professional's tacit approach into structured assets. (2)~Authorization, with autonomy escalation gated by explicit human approval. (3)~Continuous alignment, correcting drift at runtime and converting each correction into owned preference data.

We instantiate this framework as an inference-time control plane. We mathematically model the quality framework and discuss policies and techniques designed to raise quality. We apply the framework to an open professional corpus, and we show how catching data drift and applying a different technique at runtime recovers degraded quality dimensions under traffic shift.

The implication extends beyond any single application. We believe these three pillars, stitched together as a system, form a safer and more viable path to agentic systems that can scale without sacrificing trust.
\end{abstract}

\section{Introduction}
Professional adoption of agentic AI is constrained less by raw model capability than by governance. Organizations delegate autonomy to software routinely, but rarely with the three things responsible delegation requires: a way to capture the tacit methodology that defines competent work in their setting, a record of who authorized what, and continuous alignment as conditions shift. Tool-centric copilots treat each inference as stateless and accumulate no durable account of a professional's judgment. Autonomy-maximizing agents act broadly before their reliability in a specific setting has been established. Both leave the organization exposed.

We contribute a two-part answer. The Digital Apprentice is a conceptual framework for human-directed development in which an agent's authority over a skill expands only as it demonstrates competence on that skill and a human authorizes the expansion. ADAPT (Adaptive Data Augmentation and Preference Tuning) is the inference-time control plane that operationalizes the framework: it runs multiple policies, measures output quality along several dimensions, and records each human correction as a reusable preference signal that stays within the organization's environment. In this design, inference becomes a record-generating event. Each judgment is externalized, isolated to the tenant that produced it, and available for in-context steering or, when warranted, model updating. The conceptual framework specifies what autonomy looks like at each tier; the control plane specifies how that autonomy is measured, maintained, and reversed. This separation lets the framework be assessed on its own terms, with ADAPT as one implementation rather than the definition. Governance does not stop at policy; it enforces policy through data-driven decisions grounded in the professional's method.

Taxonomies describe what autonomy looks like; our contribution is the machinery for moving between levels: what evidence justifies an increase in autonomy, who must authorize it, and how the system detects when autonomy should be withdrawn.

\section{Graduated Autonomy Framework}

\subsection{Per-Skill State Machine}
Autonomy in the Digital Apprentice is a per-skill property that the system holds at a given time, represented as a finite state machine (Table~\ref{tab:autonomy}) rather than a fixed benchmark tier. The agent cannot promote itself. Promotion from one tier to the next requires both empirical evidence of competence on that skill and a recorded human authorization. Demotion is asymmetric: the system rolls a skill back automatically when quality degrades, without waiting for human action.

This per-skill, role-indexed structure is consistent with autonomy-level taxonomies that index control by the role a human retains over the agent \citep{feng2025,beer2014}. The difference is that those taxonomies describe the levels; we specify the transition conditions between them.

\begin{table}[t]
\centering
\caption{Per-skill autonomy tiers.}
\label{tab:autonomy}
\footnotesize
\begin{tabularx}{\columnwidth}{@{}>{\bfseries}l >{\raggedright\arraybackslash}X @{}>{\raggedright\arraybackslash}X @{}>{\raggedright\arraybackslash}X@{}}
\toprule
\textbf{State} & \textbf{What the agent may do} & \textbf{Human role} & \textbf{External effects} \\
\midrule
Pre-L0 & Observe-only: ingest human--human data; form hypotheses; \emph{no outputs} & Director validates inferences in guided dialogue. & None \\
L0 Sandbox & Generate isolated draft outputs for one authorized skill. & Supervisor: reviews \textbf{100\%} of outputs before they take effect. & Drafts only; no execution \\
L1 Draft & Produce candidate outputs routed toward execution. & Endorser: approves outputs before they have external effects. & After approval \\
L2 Autonomous & Execute the skill independently under continuous monitoring. & Delegator: intervenes only when exceptions arise. & Full auth; drift checks active \\
\bottomrule
\end{tabularx}
\end{table}

Entry is Pre-L0 (Observe-only). Transition to L0 requires only explicit human authorization after an observation period ($N$ sessions); the graduated promotion criteria (Eq.~\ref{eq:graduate}) apply to all subsequent tier transitions.

\subsection{Graduation Mathematics}
Let $W_t$ denote the $t$-th evaluation window of $N$ consecutive outputs for a single skill. Promotion checks use non-overlapping windows; rolling windows may be used for runtime monitoring. Let $c(x) = 1$ if output $x$ required a human correction and 0 otherwise. The per-window correction rate is:
\begin{equation}
\rho(W) = \frac{1}{|W|} \sum_{x \in W} c(x)
\label{eq:correction_rate}
\end{equation}
When a validated quality scorer $Q(x)$ is available, with acceptability threshold $\tau$, a proportion $p$ of outputs in the window must clear the threshold:
\begin{equation}
\frac{1}{|W|} \sum_{x \in W} \mathbf{1}[Q(x) \ge \tau] \ge p
\label{eq:quality_gate}
\end{equation}
Graduation from tier $L_i$ to $L_{i+1}$ requires three conditions plus authorization. \textbf{(C1) Net improvement:} the correction rate in the current window is lower than $k$ windows earlier ($\rho(W_t) < \rho(W_{t-k})$; we use $k{=}3$), accommodating normal variance while detecting genuine improvement. \textbf{(C2) Low residual correction:} $\rho(W_t) \le \tau_{\mathrm{corr}}$. \textbf{(Q1) Scorer gate:} Eq.~\ref{eq:quality_gate} holds for the current window.
Let $H_{\mathrm{auth}} \in \{0,1\}$ record an explicit human authorization event. Graduation is then:
\begin{equation}
\label{eq:graduate}
\begin{aligned}
& \mathrm{Graduate}(L_i \!\to\! L_{i+1}) \iff (C1 \land C2) \\
& \quad \land (\neg\textsc{scorer}\lor Q1) \land (H_{\mathrm{auth}}{=}1)
\end{aligned}
\end{equation}
The correction-rate gate has a known vulnerability that we address directly: a low correction rate can indicate either genuine competence or a reviewer who has stopped checking carefully. This is the automation-complacency failure documented in the human-factors literature \citep{parasuraman2010}. We treat reviewer engagement as a monitored quantity rather than an assumption. A skill is demoted automatically when $\rho(W) > \rho_{\mathrm{demote}}$ or out-of-distribution uncertainty exceeds $u_{\max}$; we use a rate-based trigger so ordinary output noise at L2 does not cause oscillation. Re-escalation requires satisfying Eq.~\ref{eq:graduate} again with new $H_{\mathrm{auth}}$.

\subsection{Two-Phase Learning}
Learning proceeds in two phases that differ in speed and reversibility. In Phase 1, human corrections populate preference pairs that serve as tenant-isolated decision memory, retrieved at inference time to steer outputs immediately. Phase 1 acts as an immediate safety buffer: steering based on recent corrections adapts within hours, while the underlying model remains unchanged. This phase is fully reversible and requires no update to the generator. In Phase 2, once the volume and statistical significance of accumulated preferences cross set thresholds, the pairs are exported to a model-updating step such as supervised fine-tuning or direct preference optimization \citep{ouyang2022,rafailov2023,christiano2017}. Determining domain-specific thresholds for Phase 2 activation is left to future work. Quality regressions on the multidimensional scorer reject a bad update before it is committed. Every update is traceable to a specific human correction or an observed human decision, which is the property that distinguishes this loop from open-ended self-improvement.

\section{ADAPT: Inference-Time Control Plane}
ADAPT operationalizes the Digital Apprentice as continuous-learning infrastructure positioned between an organization's orchestration layer and its model providers. Four components form one loop: (1)~asset synthesis (methodology, style, and authority exemplars, after professional validation); (2)~multi-policy inference (RAG, methodology-conditioned generation, best-of-$N$, and diversity-gated fusion); (3)~quality telemetry (vector scoring across named dimensions, Table~\ref{tab:radar}); (4)~preference emission (weighted pairs from human corrections and policy comparisons). These draw from classical ML (RAG, best-of-$N$ sampling) as well as control-plane techniques we introduce (methodology-conditioned generation, diversity-gated fusion).

Each inference event follows a four-step pipeline. \textbf{Branch:} given a prompt $x$, a branch policy (e.g., best-of-$N$ sampling with different model temperatures, RAG retrieval top-$k$ settings, or different prompt framings such as ``advise'' vs.~``draft'' vs.~``challenge'') produces candidate outputs $Y_N = \{y^{(1)}, \ldots, y^{(N)}\}$. \textbf{Score:} each candidate receives a radar vector $\mathbf{r}(x,y)$ (Table~\ref{tab:radar}). \textbf{Triage:} a pluggable scorer (LLM-as-judge rubric, trained reward model, or embedding centroid over approved exemplars) computes $R(y) = \mathrm{agg}(\mathbf{r})$ and ranks candidates; the system presents the highest-scoring $y^+$ to the directing professional for validation or correction, and treats the remainder as $y^-$. \textbf{Emit:} every comparison becomes a preference tuple. A rejected candidate forms an automatic tuple $(x, y^+, y^-, w_{\mathrm{auto}})$ with $w_{\mathrm{auto}} \in [0.2, 0.5]$ and \texttt{policy\_comparison} provenance. A human correction forms a tuple with $w_{\mathrm{human}} = 1.0$ and \texttt{human\_correction} provenance. These records are tenant-isolated decision memory for in-context steering or optional model updating. The pilot used $w_{\mathrm{auto}} = 0.35$.

\subsection{Methodology Quality Rubric}
We score professional quality as a six-dimensional vector rather than a single opaque number, so that drift in one dimension is visible even when others hold. The rubric dimensions were defined during structured onboarding with the directing professional, reflecting the qualities that distinguish competent from expert work in that practice. The framework supports any $d \ge 1$; we instantiate $d = 6$ here (Table~\ref{tab:radar}).

\begin{table}[t]
\centering
\caption{Quality rubric dimensions ($d = 6$).}
\label{tab:radar}
\footnotesize
\begin{tabularx}{\columnwidth}{@{}l X@{}}
\toprule
\textbf{Dimension} & \textbf{High score means} \\
\midrule
Methodology Fit & Reflects the expert's decision style, not generic advice \\
Voice / Style Fit & Output reads as the expert's apprentice \\
Grounding & Uses approved source concepts without fabrication \\
Actionability & Gives a clear, realistic next step \\
Context Sensitivity & Captures the nuance of the stated concern \\
Safety Boundary & Non-coercive, non-diagnostic, appropriately scoped \\
\bottomrule
\end{tabularx}
\end{table}

\subsection{Drift, Policy Switching, and Recalibration}
Drift is multidimensional. When incoming requests move into areas adjacent to the onboarding distribution (the same profession, but unseen topics or cases), methodology dimensions may remain strong while operational dimensions (actionability, context sensitivity, safety boundary) deform. The control plane detects this shift quantitatively: rolling-window radar telemetry on each output reveals dimension-specific score deformation (e.g., falling grounding or actionability while methodology fit holds). When per-dimension or mean scores cross calibrated thresholds, a localized regression is flagged and policy switching is triggered. The control plane distinguishes three causes: human methodology evolved (accelerate observation and incorporate new exemplars), agent regressed (rollback and increase review frequency), or evaluation criteria shifted (revise rubric definitions and re-baseline historical telemetry). Each cause requires a different response. When localized degradation is detected, ADAPT switches policy at runtime rather than serving a statically ``optimal'' onboarding policy. The control plane applies techniques from a broader policy repertoire. Classical mutual information measures pairwise statistical dependence between candidates; we instead use a lightweight alternative based on dispersion in quality-score space. For a shortlisted subset $S \subseteq Y_N$, typically the top-$k$ candidates by mean score, generated under different framings, we compute diversity as the mean pairwise Euclidean distance between radar vectors:
\begin{equation}
\Delta_f(S) = \frac{2}{|S|(|S|-1)} \sum_{1 \le i < j \le |S|} \left\lVert \mathbf{r}\bigl(y^{(i)}\bigr) - \mathbf{r}\bigl(y^{(j)}\bigr) \right\rVert_{2}
\label{eq:framing_diversity}
\end{equation}
where $\mathbf{r}(y^{(i)})$ is the six-dimensional radar vector for candidate $y^{(i)}$. Low $\Delta_f$ indicates that candidates collapse onto the same quality profile despite different framings; in this case synthesis is skipped. High $\Delta_f$ indicates candidates occupy distinct points in the quality-space, justifying fusion to recover dimensions where no single candidate is strong. This fusion may combine outputs with complementary score profiles via the fusion operator $\mathcal{F}$, which synthesizes a composite answer:
\begin{equation}
y_{\mathrm{fuse}} = \mathcal{F}\big(x, S\big) \quad \text{when } \Delta_f(S) \ge \delta
\label{eq:fuse_rule}
\end{equation}
where $\delta$ is a tunable threshold. Because radar dimensions are normalized to $[0,1]$, $\Delta_f$ is bounded by $\sqrt{d}$ (at most $\approx 2.45$ for $d=6$); $\delta$ is calibrated on validation traffic. This dispersion serves as a lightweight diversity gate for inference-time fusion: it reuses radar scores already computed by the control plane and requires no additional embedding model. At runtime, the control plane may apply, skip, or switch policies entirely when drift is detected; the diversity metric is a component, not the entire strategy. When $y_{\mathrm{fuse}}$ improves degraded dimensions without collapsing strong ones relative to the single-best candidate, a fusion tuple is emitted, recovering operational dimensions without sacrificing strong methodology scores.

\section{Related Work}

Human-in-the-loop and human-on-the-loop designs either scale poorly or intervene too late for agentic workloads \citep{wu2022}. Reinforcement learning from human feedback and direct preference optimization align a model to aggregated preferences \citep{christiano2017,ouyang2022,rafailov2023}, but they align to a population rather than to a specific directing professional's methodology, and they do not by themselves detect drift during observation or action. We expect this gap to widen as deployments move beyond text into multimodal sensing. Best-of-$N$ methods discard rejected candidates rather than retaining them as decision memory \citep{liao2026}. Autonomy taxonomies assign tiers but leave transition machinery unspecified \citep{feng2025,beer2014}; governance frameworks for agentic AI remain similarly high-level \citep{imda2026}. Corrigibility work complements our authorization gate \citep{nayebi2025}. Our contribution is the integration pattern: each inference produces a durable, tenant-isolated judgment record governing immediate steering and optional model update.

\section{Proof-of-Concept}
We instantiate ADAPT on an openly available professional-methodology corpus to illustrate the mechanisms, not to establish a general result. The setup uses $40$ to $60$ prompts per arm, a Qwen model as generator, and a Gemma model as LLM-as-judge evaluator over the six-dimensional rubric, accessed through OpenRouter \citep{openrouter2024}. Reported means are the arithmetic average of the six rubric dimensions across all prompts in the evaluation arm. Branch-and-triage does most of the narrowing work: the judge scores and ranks candidates; the directing professional then validates or corrects the presented output. Judge-ranked comparisons emit automatic preference tuples ($w_{\mathrm{auto}}{=}0.35$); human acceptance or correction emits full-weight tuples ($w_{\mathrm{human}}{=}1.0$). Figures~\ref{fig:onboarding} and~\ref{fig:drift} show triage-stage radar profiles (judge-measured, pre-human validation) under each policy arm.

\noindent\textbf{Two-arm pilot.} Arm~A measures onboarding quality by comparing corpus-only retrieval (no methodology assets) to an onboarding-guided policy. Arm~B measures runtime drift recovery by shifting live traffic to a new topic area and switching from the onboarding policy to diversity-gated fusion over retrieval candidates.

\noindent\textbf{Before and after structured onboarding.} With the corpus and plain retrieval only, the mean triage-stage score is 0.717. Best-of-$N$ sampling raises it to 0.780, and diversity-gated fusion to 0.803. After structured onboarding converts raw artifacts into methodology and style assets, the mean rises to 0.957 at triage (Figure~\ref{fig:onboarding}), before the human validates the presented output.

\noindent\textbf{Under runtime drift.} When live requests differ from those seen during onboarding, the onboarding-selected policy retains strong methodology, voice, and grounding at triage, but the mean falls to 0.930, with actionability at 0.770 and safety boundary at 0.870. Switching to diversity-gated recalibration restores actionability to 0.905 and the triage mean to 0.957 (Figure~\ref{fig:drift}); human validation of the final output remains the authority step that emits preference data.

\noindent\textbf{Limitations of this evidence.} The radar profiles are triage-stage, judge-measured telemetry on one corpus \citep{zheng2023}, illustrating the measurement-and-switching loop rather than post-human ground truth. The pilot does not report inter-rater agreement, confidence intervals, or significance testing. The intended production path is branch-and-triage first, human validation second, preference emission on every human decision.

These results support a systems claim: professional AI quality is a runtime variable, measurable, improvable, monitorable, and convertible to organization-owned preference data through the inference control plane.

\clearpage
\onecolumn
\begin{paracol}{2}
\section{Limitations and Risks}

\noindent\textbf{Tacit knowledge as an underdetermined inverse problem.} Much of what experts know cannot be fully articulated \citep{polanyi1966,nonaka1995}, and situated-action work argues that plans cannot be read off from observed behavior (\citeauthor{suchman1987}, \citeyear{suchman1987}, \citeyear{suchman2007}). Recovering a professional's methodology from observation is an ill-posed inverse mapping: many world models are consistent with any finite observation set, and any single modality captures only a compressed projection. Our current work addresses the tractable slice: text-mediated interaction as a proxy for professional judgment, where the inference space is constrained enough to yield learnable decision regularities.

\noindent\textbf{Consent and confidentiality of observed interaction.} Recording human-to-human professional interaction, for example coaching sessions or clinical exchanges, raises consent, confidentiality, and data-protection obligations, and in some jurisdictions implicates obligations under the EU AI Act \citep{euaiact2024} and proposed civil-liability rules \citep{euaidirective2022}. Any deployment must establish the lawful basis for observation and the rights of the observed parties before the Pre-L0 stage begins. We flag this as a design precondition, not an afterthought.

\noindent\textbf{Trust.} Promotion reads low correction rate as competence, creating complacency risk \citep{parasuraman2010,dietvorst2015,devissier2018}. Reviewer engagement must be monitored; we recommend periodic seeded checks.

\vspace{2pt}
\section{Conclusion}

The Digital Apprentice and ADAPT treat governance as an in-deployment property rather than a fixed configuration. Observational learning grounds capability in demonstrated practice; graduation gates and authorization lineage govern escalation; multidimensional measurement and runtime policy switching maintain alignment as conditions shift; and accumulated preference pairs form a durable, organization-owned improvement substrate. We suggest that continuous, human-grounded learning infrastructure is a productive direction for human-directed agentic AI in high-stakes settings. The measure of success is not how autonomous AI becomes, but how much further human expertise can reach.

\switchcolumn

\begin{center}
\includegraphics[width=\linewidth,height=0.34\textheight,keepaspectratio]{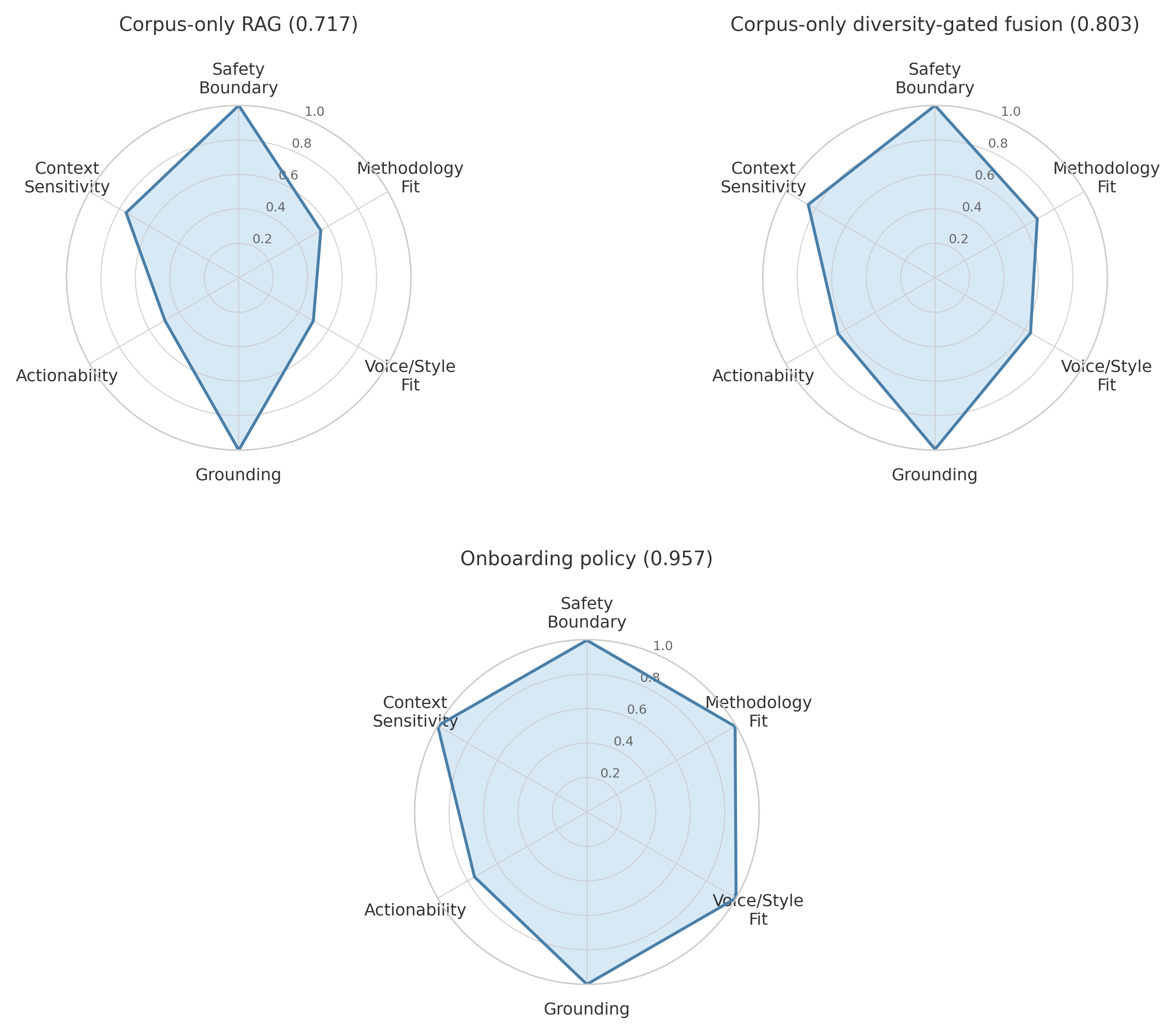}
\captionof{figure}{Before/after structured onboarding ($n{=}40$ to $60$): corpus-only RAG (0.717), corpus-only diversity-gated fusion (0.803), and onboarding-guided policy (0.957).}
\label{fig:onboarding}
\end{center}

\vspace{6pt}
\begin{center}
\includegraphics[width=\linewidth,height=0.36\textheight,keepaspectratio]{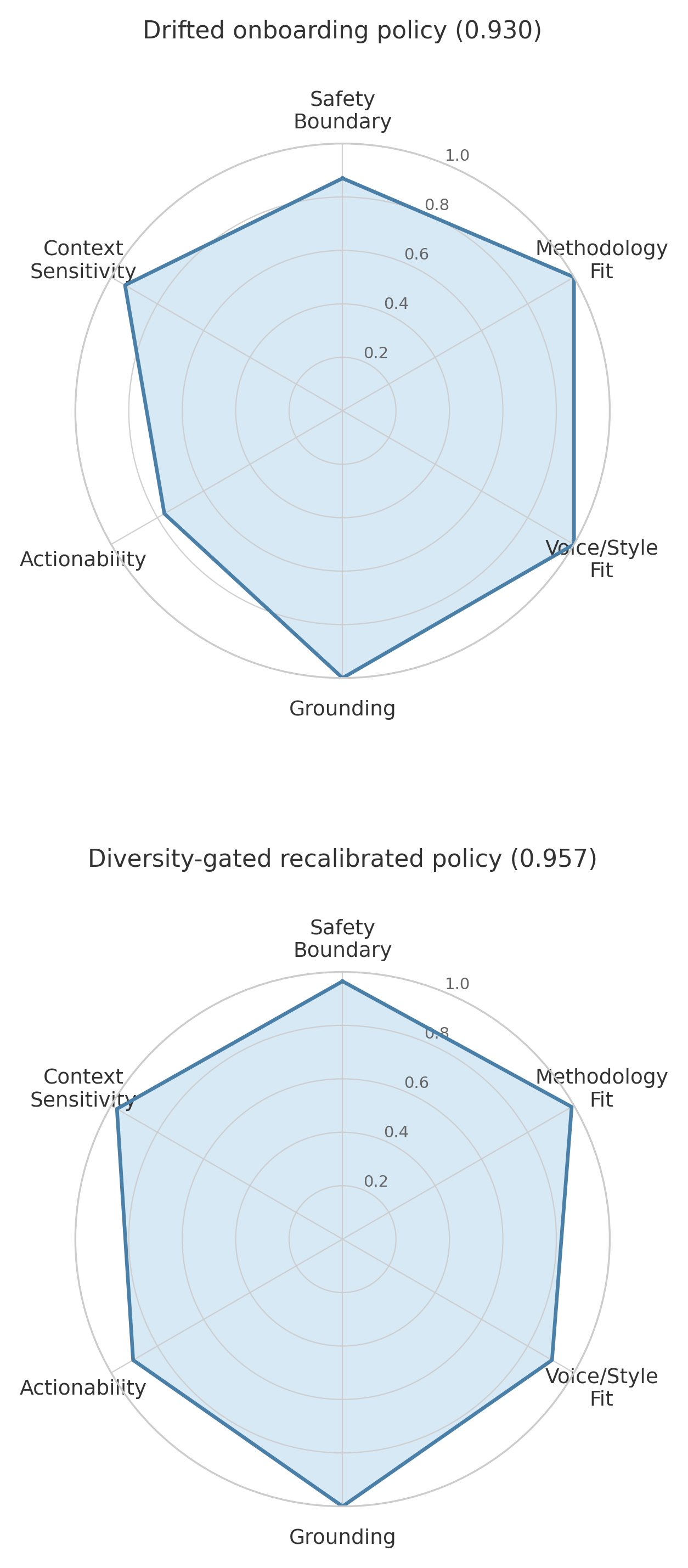}
\captionof{figure}{Before/after runtime drift and policy recalibration (Arm~B: $n{=}40$): drifted onboarding policy (0.930) and diversity-gated recalibrated policy (0.957).}
\label{fig:drift}
\end{center}

\end{paracol}

\twocolumn
\section*{References}
\renewcommand{\refname}{}

\end{document}